\documentclass[letterpaper, 10 pt, conference]{ieeeconf}  
\IEEEoverridecommandlockouts                         
\usepackage[T1]{fontenc}
\usepackage{aecompl}
\usepackage{epstopdf}
\usepackage{cite}
\usepackage{amsmath,amssymb,amsfonts}
\usepackage{algorithmic}
\usepackage{algorithm}
\usepackage{graphicx}
\usepackage{textcomp}
\usepackage{xcolor}
\usepackage{float}
\usepackage{graphicx}
\usepackage{picinpar}
\usepackage{babel}
\usepackage{amsmath}
\usepackage{url}
\usepackage[latin1]{inputenc}
\usepackage{colortbl}
\usepackage{soul}
\usepackage{multirow}
\usepackage{pifont}
\usepackage{color}
\usepackage{alltt}
\usepackage[hidelinks]{hyperref}
\usepackage{enumerate}
\usepackage{siunitx}
\usepackage{breakurl}
\usepackage{epstopdf}
\usepackage{pbox}
\usepackage{authblk}
\usepackage{babel}
\usepackage{mathrsfs,balance,subfigure}
\usepackage{tcolorbox}

\title{\LARGE \bf Model-Based Actor-Critic with Chance Constraint for Stochastic System}

\author{Baiyu Peng$^{1}$, Yao Mu$^{1}$, Yang Guan$^{1}$, Shengbo Eben Li$^{1*}$, Yuming Yin$^{1}$, Jianyu Chen$^{2}$% <-this % stops a space
\thanks{This study is supported by Tsinghua-Toyota Joint Research Institute Cross-discipline Program and Xilinx. }% <-this % stops a space
\thanks{$^{1}$State Key Laboratory of Automotive Safety and Energy, School of Vehicle and Mobility, Tsinghua University, Beijing, China  ${}^{*}$ Corresponding author {\tt\small lishbo@tsinghua.edu.cn}}%
\thanks{$^{2}$Institute for Interdisciplinary Information Sciences, Tsinghua University, Beijing, China {\tt\small jianyuchen2020@163.com}}%
}

\begin{document}
\maketitle
\thispagestyle{empty}
\pagestyle{empty}

\begin{abstract}
Safety is essential for reinforcement learning (RL) applied in real-world situations. Chance constraints are suitable to represent the safety requirements in stochastic systems. Previous chance-constrained RL methods usually have a low convergence rate, or only learn a conservative policy. In this paper, we propose a model-based chance constrained actor-critic (CCAC) algorithm which can efficiently learn a safe and non-conservative policy. Different from existing methods that optimize a conservative lower bound, CCAC directly solves the original chance constrained problems, where the objective function and safe probability is simultaneously optimized with adaptive weights. In order to improve the convergence rate, CCAC utilizes the gradient of dynamic model to accelerate policy optimization. The effectiveness of CCAC is demonstrated by a stochastic car-following task. Experiments indicate that compared with previous RL methods, CCAC improves the performance while guaranteeing safety, with a five times faster convergence rate. It also has 100 times higher online computation efficiency than traditional safety techniques such as stochastic model predictive control.  
\end{abstract}

% ￥这里要突出减少保守性吗，在还没有提到其他方法有保守性的时候
% By leveraging Boole's inequality approximation, the joint chance constraint is converted to a constraint of an additional cost function \cite{Ono2015ChanceconstrainedDP} and the exterior point method is adopted to transfer the cost-constrained problem to an unconstrained problem. We solve the unconstrained problem by model-based method which utilize the gradient of state transition as a guidance to policy optimization. To decrease the conservatism, the weight of the constraint is adaptively adjusted according to the safe probability, which can be estimated by sampling large numbers of trajectories thanks to the access of the dynamic model.

\section{Introduction}
Recent advances in deep reinforcement learning (RL) have demonstrated state-of-the-art performance on a broad set of tasks, including Atari games \cite{Mnih2015HumanlevelCT}, StarCraft \cite{Vinyals2019GrandmasterLI} and Go \cite{Silver2017MasteringTG}. However, these works do not consider safety since they are usually applied in virtual games. In many real-world tasks such as autonomous driving and unmanned aerial vehicles, the agent should follow some safety rules besides achieving excellent performance. For instance, an autonomous car driving on the highway, while optimizing its velocity, must not take actions that may cause a crash with the surrounding car. Usually, it is nontrivial to learn a driving policy that is both efficient and safe.  \cite{Li2015MechanismOV}. 
%加一个例子

% The safety concept, traditionally, has taken mainly three forms in the safe RL community, which contains worst-case performance criterion, risk-sensitive criterion, and constrained criterion \cite{Garcia2015ACS}. 分别讨论worst case 和 risk-sensitiv的缺陷. On the other hand, it is more natural to define safety in a constrained criterion这篇文章考虑约束准则, i.e., the agents achieve the main goal while try not violating the state constraints. 

The safety consideration has taken different forms in the safe RL community \cite{Garcia2015ACS,DulacArnold2019ChallengesOR}. Tamar (2013) treated safety in a robust view and optimized the worst-case performance of the agent \cite{Tamar2013VarianceAA}. Chow (2017) used value-at-risk as a metric of safety and a policy was regarded safe if its value-at-risk was high enough \cite{Chow2017RiskConstrainedRL}. Recently, many researchers also cast safety in the context of Constrained MDPs, where the cumulative cost was constrained below a given threshold \cite{achiam2017constrained, tessler2018reward,yang2019projection}. However, these criteria all focus on reward-related or cost-related measures, and they still lack a direct connection with safety \cite{geibel2005risk}. In other words, given a learned policy with a certain value-at-risk, it is still hard to evaluate how safe the policy is. Indeed, an explicit safety constraint is preferred in real-world applications \cite{mayne2000constrained, Duan2019DeepAD}. In this work, we aim to build a safe policy optimization framework, which can quantitatively constrain the possibility of the control policy violating the state constraint.
%  . The state constraint widely used in control area is actually a good safety measure , so this paper considers safety as agents satisfying some state constraints, i.e., not entering some dangerous states.
It should be stressed that plenty of real-world systems are stochastic in nature, and thus the state constraint only holds in a probability form, which is quite different from the hard constraint in deterministic systems. For example, in the case of an unmanned aerial vehicle, the direction and force of wind are uncertain. Thus it can only keep balance at a high probability. Specifically, the state constraint in such a probability form is briefly called chance constraint.
% 详细说明用来处理CC的可行的方法

Strategies used to solve the chance constrained reinforcement learning problems can be roughly categorized into two approaches. The first and the most common solution is to add a fixed-weight penalty term to the reward function so as to prevent agents from entering the dangerous states \cite{Duan2020HierarchicalRL, Guan2019CentralizedCC}. Although this approach is very straightforward and simple to implement, it requires the penalty weight to  strike a balance between safety and performance correctly. Unfortunately, it is usually difficult to select an appropriate fixed-weight. Especially, a large penalty is prone to converge to sub-optimal solutions, while a small penalty is unable to satisfy the constraint \cite{tessler2018reward}. The second approach constrains the lower bound of safe probability to the required threshold, which can be solved through dynamic programming method \cite{Ono2015ChanceconstrainedDP}  or model-free primal-dual (MF-PD) method \cite{Paternain2019LearningSP,Paternain2019SafePF}. Nevertheless, the dynamic programming method only works in discrete state and action space, which can not be applied to continuous problems. The model-free primal-dual method is purely data-driven, which leads to high variance and low convergence rate. Moreover, constraining the lower bound of safe probability may produce a policy whose real safe probability is significantly higher than the required threshold, i.e, introduces large conservatism. As shown in our experiments, the learned policy achieves 99\% safe probability even when the required threshold is only 90\%, and thus influences the performance. 

To overcome the aforementioned challenges, this paper proposes a model-based algorithm named chance constrained actor-critic (CCAC). Instead of constraining the lower bound of safe probability like MF-PD, CCAC directly solves the original chance constrained problems through the exterior point methods. In order to improve the convergence rate, the gradient of dynamic model is utilized to guide policy optimization. Finally,  CCAC is compared with two RL methods and two traditional safety techniques such as stochastic model predictive control to demonstrate its superior performance. The contributions of this paper are as follows,
\begin{enumerate}
\item a direct approach to solve the chance constrained problems, rather than indirectly solving by constraining the lower bound of safe probability.

\item a model-based framework of policy optimization for chance constrained problems, where the gradient of the dynamic model is used to accelerate training process.
\end{enumerate}
 
The rest of this paper is organized as follows. The chance constrained RL problem is formulated in Section \ref{sec:Preliminary}. The  CCAC algorithm is proposed in Section \ref{sec:chance constrained Actor-Critic Algorithm}. The effectiveness of CCAC is illustrated by a stochastic car-following task in Section \ref{sec:Numerical Experiment}. Section \ref{sec:Conclusion}  concludes this paper.

\section{Preliminary}
\label{sec:Preliminary}
For a discrete-time stochastic system, the dynamics with the chance constraint is mathematically described as:
\begin{equation}
\begin{aligned}
&s_{t+1}=f(s_{t}, a_{t}, \xi_{t}), \quad \xi_{t}\sim p(\xi_{t}),\\
&{\rm Pr}\left\{ \bigcap_{i=1}^{N} \left[h\left(s_{t+i}\right)<0 \right]\right\}\ge1-\delta
\end{aligned}
\end{equation}
%h\left(x_{t+i}\right)<0
where $t$ is the current time step, $s_{t}\in{\mathcal{S}}$ is the state, $a_{t}\in {\mathcal{A}} $ is the action, $f(\cdot,\cdot,\cdot)$ is the environmental dynamics, $\xi_{t} \in {\mathbb{R}}^{n}$ is the uncertainty following an independent and identical distribution $p(\xi_{t})$, $h(\cdot)$ is the state constraint function and the set $\left\{s\mid h(s)<0 \right\}$ defines a safe region in which the agent should remain. We do not make assumptions about the form of $f(\cdot,\cdot,\cdot)$ and $h(\cdot)$, i.e., they can be linear or nonlinear. The safety constraint takes form of a joint chance constraint with $1-\delta$ as the required threshold. Such a form is extensively used in stochastic systems control \cite{Mesbah2016StochasticMP}. Intuitively, it can be interpreted as the probability of agent staying within a safe region $\left\{s\mid h(s)<0 \right\}$ over the horizon $N$ is at least $1-\delta$. For simplicity, we only consider one constraint.

The objective of chance constrained RL problems is to maximize the expectation of cumulative reward $J_r$, while constraining the safe probability $p_s$:
\begin{equation}
\begin{aligned}
&\max _{\pi} J_{r}\left(\pi\right)=\mathbb{E}_{s_0,\xi}\left\{\sum_{t=0}^{\infty} \gamma^{t} r\left(s_{t}, a_{t}\right)\right\}\\
&\text { s.t. }p_s(\pi)={\rm Pr}\left\{ \bigcap_{t=1}^{N} \left[h\left(s_{t}\right)<0\right]\right\}\ge1-\delta
\label{CCRL problem}
\end{aligned}
\end{equation}
where $\pi$ is the policy, $r(\cdot,\cdot)$ is the reward function, $0<\gamma<1$ is the discounting factor and ${\mathbb{E}_{s_0,\xi}}(\cdot)$ is the expectation w.r.t. the initial state $s_0$ and uncertainty $\xi$. Specifically, the policy is a deterministic mapping from state space ${\mathcal{S}}$ to action space ${\mathcal{A}}$ with parameters $\theta$ : $ a_{t}=\pi(s_{t};\theta)$.

The joint chance constraint in \eqref{CCRL problem} is generally nonconvex and intractable \cite{Bertsimas2011TheoryAA}. Therefore, previous methods like model-free primal-dual (MF-PD) usually solve the chance constraint indirectly, i.e, derive a lower bound of the joint probability $p_s$ through the Boole's inequality and turn to constrain this lower bound \cite{Ono2015ChanceconstrainedDP, Paternain2019LearningSP}. 
% \subsection{Reformulation of Chance Constraint via Boole's Approximation}
% The chance constraint is hard to handle in RL setting. Especially, the key difficulty lies in two problems: (1) how to estimate the safe probability, and (2) how to calculate an ascend direction of the safe probability. 
% %以上内容待考虑
% We first reform chance constraint via Boole's approximation \cite{Ono2015ChanceconstrainedDP}. 
More specifically, a cost function $c(s_t,a_t,s_{t+1})$ and the expected cumulative cost $J_c$ are defined as:
\begin{equation} c(s_t,a_t,s_{t+1})=
\begin{cases}
0& \text{$h(s_{t+1})<0$} \\
1& \text{$h(s_{t+1})\ge0$}
\end{cases}
\end{equation}

\begin{equation}
J_{c}(\pi)={\mathbb{E}_{s_0,\xi}}\left\{\sum_{t=0}^{N-1}c(s_{t},a_{t},s_{t+1})\right\}
\label{Jc}
\end{equation}

% The expectation of cost function represents the single-step unsafe probability
% \begin{equation} 
% {\rm Pr}\left\{h\left(x'\right)\ge 0\right\}={\mathbb{E}}_{x'}\left\{c(x,u,x')\right\}
% \label{expected_cost}
% \end{equation} 
% Now consider the chance constraint in \eqref{CCRL problem}. Given Boole's inequality and  \eqref{expected_cost}, we have
Given the Boole's inequality, a lower bound of $p_s$ is derived as:
\begin{equation}
\begin{aligned}
p_s(\pi)&={\rm Pr}\left\{ \bigcap_{t=1}^{N} \left[h\left(s_{t}\right)<0\right]\right\}\\
&\geq1-\sum_{t=1}^{N}{\rm Pr}\left\{h\left(s_{t}\right)\ge0 \right\}\\
&=1-J_c(\pi)
\label{boole}
\end{aligned}
\end{equation}

% Thus, the chance constraint can be converted to the constraint on expected cumulative cost $J_{c}(\pi)$
% \begin{equation}
% J_{c}(\pi)={\mathbb{E}}\left\{\sum_{t=0}^{N-1}c(x_{t},u_{t},x_{t+1})\right\}\leq\delta
% \label{Jc}
% \end{equation}
% Consequently, the original problem \eqref{CCRL problem} is replaced by a new constrained RL problem  \eqref{CMDP problem}

% \begin{equation}
% \begin{aligned}
% &\max _{\pi} J_{r}\left(\pi\right)={\mathbb{E}}\left\{\sum_{t=0}^{\infty} \gamma^{t} r\left(x_{t}, u_{t}\right)\right\}\\
% &\text { s.t. } J_{c}\left(\pi\right)={\mathbb{E}}\left\{\sum_{t=0}^{N-1}c(x_{t},u_{t},x_{t+1})\right\}\leq\delta
% \label{CMDP problem}
% \end{aligned}
% \end{equation}
The original chance constraint in \eqref{CCRL problem} is indirectly imposed by constraining its lower bound:
\begin{equation}
\label{J_c_constraint}
    J_c(\pi)\le\delta
\end{equation}
Many previous methods (e.g. MF-PD) adopt above reformation because $J_{c}$ has similar additive structure as the objective function $J_{r}$, making it easier to impose the constraints in the RL framework. However, constraining the lower bound may lead to a policy whose real safe probability is significantly higher than the required threshold, i.e., introduces conservatism, as our experiments show in section \ref{sec:Numerical Experiment}. This problem is also a main challenge of a class of existing methods.

\section{Chance Constrained Actor-Critic Algorithm}
\label{sec:chance constrained Actor-Critic Algorithm}
Different from previous methods which constrain the lower bound of safe probability, we propose a model-based approach to directly solve the original chance constrained problem \eqref{CCRL problem} with less conservatism. Besides, our method also takes use of the gradient of dynamic model to accelerate the training process.
\subsection{Constrained Policy Optimization via Exterior Point Methods}\label{exterior section}
\begin{figure}[htbp]
\centerline{\includegraphics[width=0.425\textwidth]{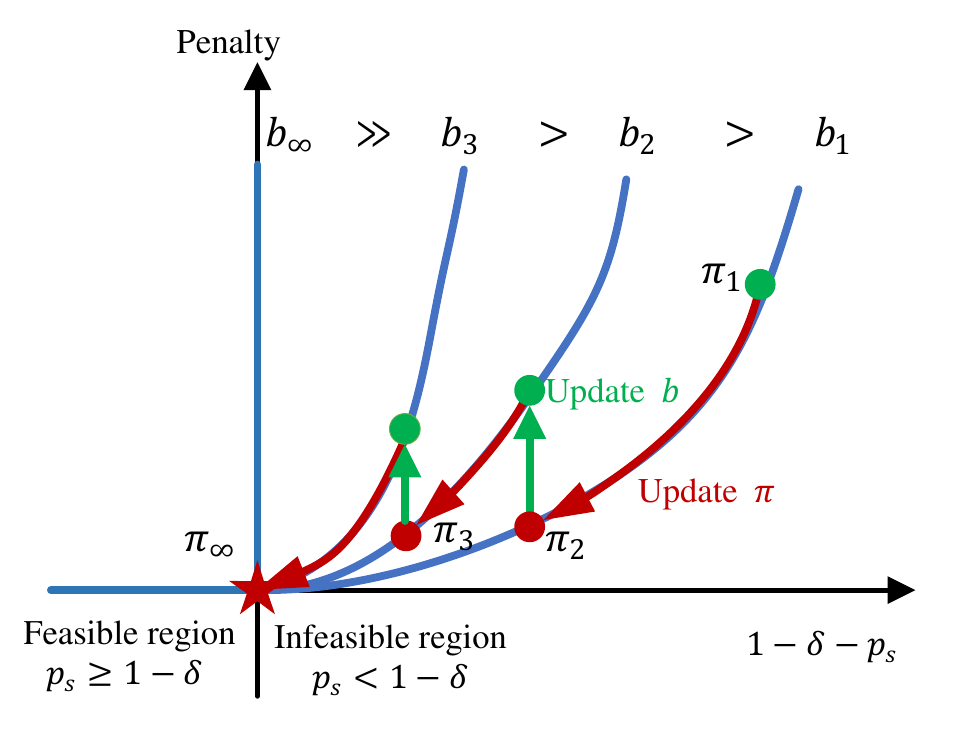}}
\caption{Exterior point methods.}
\label{fig:exterior method}
\end{figure}

The adopted approach follows the idea of exterior point methods, which are extensively used in constrained optimization area \cite{Boyd2006ConvexO}. The exterior point methods put the chance constraint into a large and increasing penalty term in the objective function in $k$-th iteration:
\begin{equation}
\max _{\pi_{k}} J_{EP}\left(\pi_{k}\right)=J_{r}\left(\pi_{k}\right) - \frac{1}{2} b_{k}\max(1-\delta-p_{s}\left(\pi_{k}\right), 0)^2
\label{exterior}
\end{equation}
where $b_{k}\gg 0$ is the penalty factor and $\left\{b_{k}\right\}$ is a given monotone increasing sequence. Intuitively, the exterior point methods penalize the violation of constraint as shown in Fig.  \ref{fig:exterior method}. As $b_{k}$ increases, the penalty will become tremendous, pushing $\pi_k$ to the feasible region. Although the intermediate policy may be infeasible, the convergent policy will be feasible. This is also the reason why it is called exterior point method.  

The chance constrained RL problem  \eqref{CCRL problem} is solved by iteratively updating $\pi_{k}$ and $b_{k}$ as shown in Fig. \ref{fig:exterior method}. However, in practice, the cost of solving $\pi_{k}$ in every iteration until convergence is computationally prohibitive, and an alternative is to replace the maximization by a gradient ascent step
\begin{equation}
\theta^{k+1} = \theta^{k} + \alpha_{\theta}\nabla_{\theta}J_{EP} \\
\end{equation}
where $\alpha_{\theta}>0$ is the learning rate of policy and the policy gradient $\nabla_{\theta}J_{EP}$ is derived as   
\begin{equation}
\begin{split}
\nabla_{\theta}J_{EP}&=
\begin{cases}
\nabla_{\theta}J_{r}& \text{$p_{s}\ge1-\delta$} \\
\nabla_{\theta}J_{r}+b_k(1-\delta-p_s)\nabla_{\theta}p_{s}& \text{$p_{s}<1-\delta$}
\end{cases}
\end{split}
\label{exterior_jc}
\end{equation}

In order to compute the above gradient, we have to obtain the current safe probability $p_s$ and its gradient $\nabla_{\theta}p_s$. Thanks to the available model, the safe probability $p_{s}$ can be easily estimated by sampling large numbers of trajectories. Specifically, we rollout $M$ trajectories with policy $\pi$. Suppose there are $m$ safe trajectories, then the safety probability is estimated by $p_s\approx \frac{m}{M}$.  Note that this rollout procedure will not impose extra computation burden since these trajectories are also necessary for the update of actor-critic as we will discuss in \ref{actor-critic}. Unfortunately, the gradient $\nabla_{\theta}p_s$ is still hard to obtain, which is also a key difficulty in solving the chance constrained problems. One possible solution is to find a substitute ascent direction to replace $\nabla_{\theta}p_s$. Inspired by the inequality $p_s\ge1-J_c$ as shown in \eqref{boole}, a decreasing $J_c$ will push $p_s$ to the ascent direction, so we replace $\nabla_{\theta}p_s$ with $-\nabla_{\theta}J_c$. Consequently, the new proxy policy gradient becomes:

%  Although the method above may obtain a feasible solution for the original chance constrained problem, it is derived based on a sufficient but not necessary condition  \eqref{boole} as in \cite{Ono2015ChanceconstrainedDP}. Consequently, conservatism will be introduced by using this sufficient condition. We propose to solve this problem by embedding a safety evaluator into the policy gradient.

%  Observe that the policy gradient  \eqref{exterior_jc} can be regarded as adaptively adjusting the relative size of gradient $\nabla_{\theta}J_{r}$ and gradient $\nabla_{\theta}J_{c}$, where the weight is decided by how much the constraint $J_c\le\delta$  in  \eqref{CMDP problem} are violated. However, the constraint $J_c\le\delta$ is only a sufficient condition of chance constraint $p_s\ge1-\delta$ in  \eqref{CCRL problem}. This mismatch between the two constraints is the cause of conservatism. To address this issue, we should adjust the weight of $\nabla_{\theta}J_{c}$ according to the satisfaction of true safe probability $p_{s}$ instead of $J_c$, i.e., replace the weight $J_{c}-\delta$ by $p_{s}-(1-\delta)$, i.e.,  The new proxy gradient becomes
 
\begin{equation}
\nabla_{\theta}J_{PR}=
\begin{cases}
\nabla_{\theta}J_{r}& \text{$p_{s}\ge1-\delta$} \\
\nabla_{\theta}J_{r}-b_k(1-\delta-p_s)\nabla_{\theta}J_{c}& \text{$p_{s}<1-\delta$}
\end{cases}
\label{exterior_p}
\end{equation}

This policy gradient can be intuitively interpreted as follows. In order to solve the chance constrained problem \eqref{CCRL problem}, we simultaneously optimize the cumulative reward and the safe probability by gradient ascent. To balance the two objectives, the weight of $\nabla_{\theta}J_{c}$ is adaptively adjusted according to the current safe probability.  We stress that CCAC is essentially different from previous methods which only constrain the lower bound. In CCAC, as long as the chance constraint $p_s\ge1-\delta$ is satisfied, the weight of $\nabla_{\theta}J_{c}$ becomes zero and the safe probability will not be optimized. While previous methods keep optimizing $p_s$ until $J_c\le\delta$, even when $p_s\ge1-\delta$ is already satisfied. That is the underlying reason why CCAC is not conservative as previous methods. Finally, in practice, since the weight $b_k(1-\delta-p_s)$ in \eqref{exterior_p} may become excessively large, we instead use the relative weights between $\nabla_{\theta}J_{r}$ and $\nabla_{\theta}J_{c}$.
%  Although the proposed method also takes use of the lower bound of $p_s$, it's essentially different from previous methods which constrain the lower bound. The main difference is that our method still searches for a policy satisfying the chance constraint in \eqref{CCRL problem}, where the lower bound only plays as a proxy gradient. While previous methods search for the policy satisfying the converted constraint \eqref{} 

%  Besides, the new gradient \eqref{exterior_p} can be also interpreted from another perspective as follows. Considering that it is hard to directly derive $\nabla_{\theta}p_{s}$, we instead take $\nabla_{\theta}J_{c}$ as a proxy of $\nabla_{\theta}p_{s}$ since  \eqref{boole} shows $1-J_{c}$ is a lower bound of $p_{s}$. Then the weight of $\nabla_{\theta}J_{c}$ is adaptively adjusted according to the current safety level.
 
\begin{figure}[htbp]
\centering 
\centerline{\includegraphics[width=0.45\textwidth]{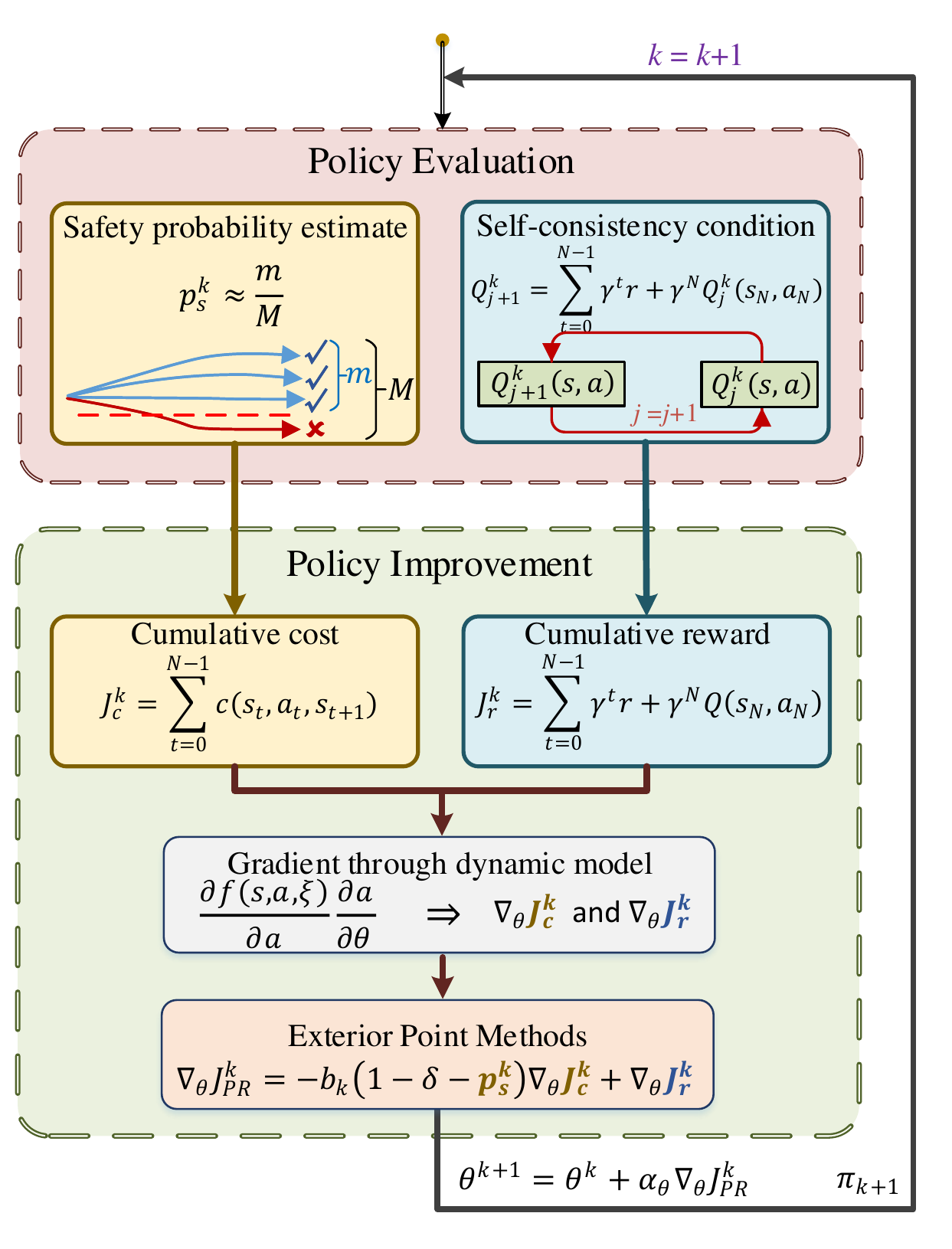}}
\caption{The framework of CCAC algorithm.}
\label{fig:CCAC algorithm}
\end{figure}

The proposed algorithm CCAC is summarized in Algorithm \ref{alg:CCAC} and Fig. \ref{fig:CCAC algorithm}.

\begin{algorithm}[!htb]
\caption{CCAC algorithm}
\label{alg:CCAC}
\begin{algorithmic}
\STATE Initialize $s_{0}$, $b_0$, $k=0$
\REPEAT
\STATE{Rollout $M$ trajectories by $N$ steps via dynamic model}
\STATE{Estimate safe probability through trajectories}
\STATE{\quad $p_s\leftarrow \frac{m}{M}$}
% \STATE{\quad $J_{c}\approx\frac{1}{M}\left\{\sum_{t=0}^{N-1}c(x_{t},u_{t},x_{t+1})\right\}$}
\STATE {Update critic according to \eqref{semi-gradient of the critic}:}
% \STATE{\quad $Q^k(x,u)={\mathbb{E}}\left\{\sum_{i=0}^{N-1} \gamma^{i} r+\gamma^NQ^k(x_N,u_N)\right\}$
\STATE{\quad  $\omega^{k+1} \leftarrow \omega^{k} + \alpha_{\omega}\nabla_{\omega}J_{Q}$ }
\STATE {Update actor according to \eqref{gradient_J_r}:}
\STATE{\quad  $\theta^{k+1} \leftarrow \theta^{k} + \alpha_{\theta}\nabla_{\theta}J_{PR}$}
\STATE{\quad  $\nabla_{\theta}J_{PR}=\nabla_{\theta}J_{r}-\text{max}(1-\delta-p_s,0)b_k\nabla_{\theta}J_{c}$ }
\STATE{Update penalty factor $b_k$}
\STATE $k\leftarrow k+1$
\UNTIL $|Q^{k+1}-Q^{k}|\le \zeta$ and $|\pi^{k+1}-\pi^{k}| \le \zeta$
\end{algorithmic}
\end{algorithm}

\subsection{Model-based Actor-Critic with Parameterized Functions}
\label{actor-critic}
In this subsection the main focus is on how to learn the policy and state-action values in the model-based actor-critic framework with parameterized functions. Importantly, the gradient of dynamic model will be utilized to attain an accurate ascent direction and thus improve the convergence rate \cite{Li2019, deisenroth2011pilco}.
For an agent behaving according to policy $\pi$, the values of the state-action pair $(s,a)$ are defined as follows:
\begin{align}
    &Q^{\pi}(s,a)={\mathbb{E}_{\xi}}\left\{\sum_{t=0}^{\infty} \gamma^{t} r\left(s_{t}, a_{t}\right)\Big|s_0=s,a_0=a\right\}
\end{align}
The expected cumulative reward $J_{r}$ can be expressed as a $N$-step form:
\begin{align}
\label{n-step}
    &J_{r}(\pi)={\mathbb{E}_{s_0,\xi}}\left\{\sum_{t=0}^{N-1}\gamma^{t} r\left(s_{t}, a_{t}\right)+\gamma^{N}Q^{\pi}(s_N,a_N)\right\}
\end{align}

For large and continuous state spaces, both value function and policy are parameterized, as shown in  \eqref{para}. The parameterized state-action value function with parameter $w$ is usually named the ``critic'', and the parameterized policy with parameter $\theta$ is named the ``actor'' \cite{Li2019}.
 \begin{equation}
     Q(s,a) \cong Q(s,a ; w), \quad a \cong \pi(s ;\theta)
     \label{para}
\end{equation}
The parameterized critic is trained by minimizing the average square error: 
\begin{equation}
\begin{aligned}
J_{Q}={\mathbb{E}}_{s_0,\xi}\left\{\frac{1}{2}\left(Q_{\text {target}}-Q(s_0,a_0;w^k)\right)^{2}\right\}
     \end{aligned}
\label{td error}
\end{equation}
where $Q_{\text {target}} = \sum_{t=0}^{N-1} \gamma^{t} r\left(s_{t}, a_{t}\right)+\gamma^{N} Q\left(s_{N},a_{N};w^k\right)$ is the $N$-step target. Note that the rollout length $N$ is equal to the horizon of chance constraint.
The semi-gradient of the critic is
 \begin{equation}
\begin{aligned}
 \nabla_{\omega}J_Q&={\mathbb{E}}_{x_0,\xi}\left\{\left(Q(s_0,a_0;w^k)-Q_{\text {target}}\right) \frac{\partial Q(s_0,a_0;w^k)}{\partial w}\right\}
\end{aligned}
\label{semi-gradient of the critic}
\end{equation}
As discussed in \ref{exterior section}, the parameterized actor aims to maximize $J_{EP}$ via gradient ascent.  The proxy gradient $\nabla_{\theta}J_{PR}$ is composed of $\nabla_{\theta}J_{r}$ and $\nabla_{\theta}J_{c}$, which are computed via backpropagation though time. Denoting $\frac{\partial s_{t}}{\partial \theta}$ as $\phi_t$,  $\frac{\partial a_{t}}{\partial \theta}$ as $\psi_t$, then $\nabla_{\theta}J_{r}$ is derived as:
\begin{equation}
\label{gradient_J_r}
\begin{split}
      \nabla_{\theta}J_{r} = &{\mathbb{E}}_{s_0,\xi}\left\{\sum_{t=0}^{N-1}\gamma^{t}\left[\frac{\partial r(s_t,a_t)}{\partial s_t}\phi_t+ \frac{\partial r(s_t,a_t)}{\partial a_t}\psi_t \right] \right. \\
     &\left. +\gamma^N \left[\frac{\partial Q(s_N,a_N)}{\partial s_N}\phi_N+\frac{\partial Q(s_N,a_N)}{\partial a_N}\psi_N     \right]\right\}
\end{split}
\end{equation}
where
\begin{equation*}
    \phi_{t+1}=\phi_{t}\frac{\partial f(s_t,a_t,\xi_t)}{\partial s_t}+\psi_{t}\frac{\partial f(s_t,a_t,\xi_t)}{\partial a_t}
\end{equation*}
with $\phi_{0}=0$, and
\begin{equation*}
    \psi_{t}=\phi_t\frac{\partial \pi(s_t;\theta)}{\partial s_t}+\nabla_{\theta}\pi(s_t;\theta)
\end{equation*}
The gradient $\nabla_{\theta}J_{c}$ can be derived similar to \eqref{gradient_J_r}. Considering that $c\left(s, a,s'\right)$ is an indicator function with zero gradient, it is replaced by the sigmoid function:  
 \begin{equation}
    c(s,a,s')=\text{sigmoid}(\eta h(s'))
\end{equation}
where $\eta>0$ is a scale factor. 
The benefits of calculating $\nabla_{\theta}J_{r}$ and  $\nabla_{\theta}J_{c}$ in the model-based framework is that the gradients of first $N$ steps' reward are computed analytically through the dynamic model. In contrast, model-free methods can not obtain these analytical gradients and thus only relies on the value function, which is usually inaccurate with high variance. In a word, the model-based framework achieves a faster convergence rate due to a more accurate gradient  \cite{deisenroth2011pilco}. The convergence and optimality of model-based actor-critic framework have been well-studied in \cite{liu2017adaptive}.

\section{Numerical Experiment}
\label{sec:Numerical Experiment}
\subsection{Experiment Setup}
\begin{figure}[htbp]
\centerline{\includegraphics[width=0.4\textwidth]{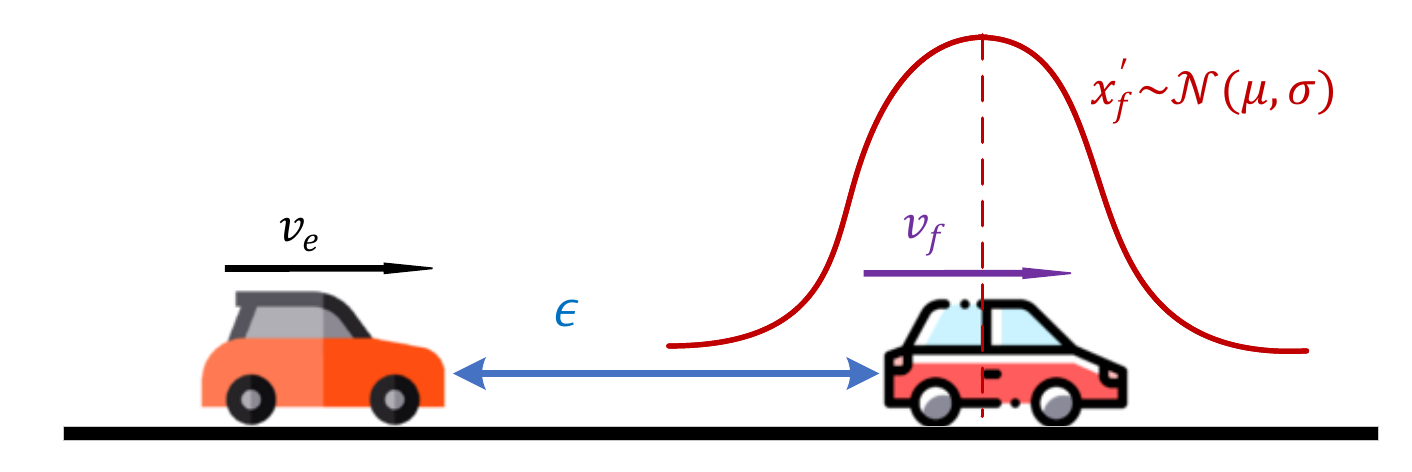}}
\caption{Car-following scenario.}
\label{fig:car-follwoing}
\end{figure}

In this section the proposed CCAC is applied to a stochastic car-following scenario as shown in Fig. \ref{fig:car-follwoing}, where the ego car expects to drive closely with the front car to reduce wind drag, while keeping a minimum gap between the two cars. Concretely, the ego car and front car follow the kinematics model, where the front car is assumed to drive at a constant speed $v_{f}$ but its location $x_{f}'$ is varying with uncertainty (e.g., due to the varying of road grade, wind drag). The minimum gap between the two cars is required to be kept at a high probability. 

The discrete-time stochastic system is described by  

\begin{equation}
\begin{aligned}
&s_{t+1}=A s_t+B a_t+D \xi_t \\
&A=\left[{\small{\begin{array}{ccc}
1 & 0 & 0 \\
0 & 1 & 0 \\
-T & T & 1
\end{array}}}\right]\\
&B=[T,0,0]^{\top}, \quad D=[0,0,T]^{\top}
\end{aligned}
\end{equation}

\begin{figure*}[hbt]
    \centering
    \subfigure[Discounted cumulative reward under $90.0\%$ threshold]{
        \includegraphics[width=0.38\textwidth]{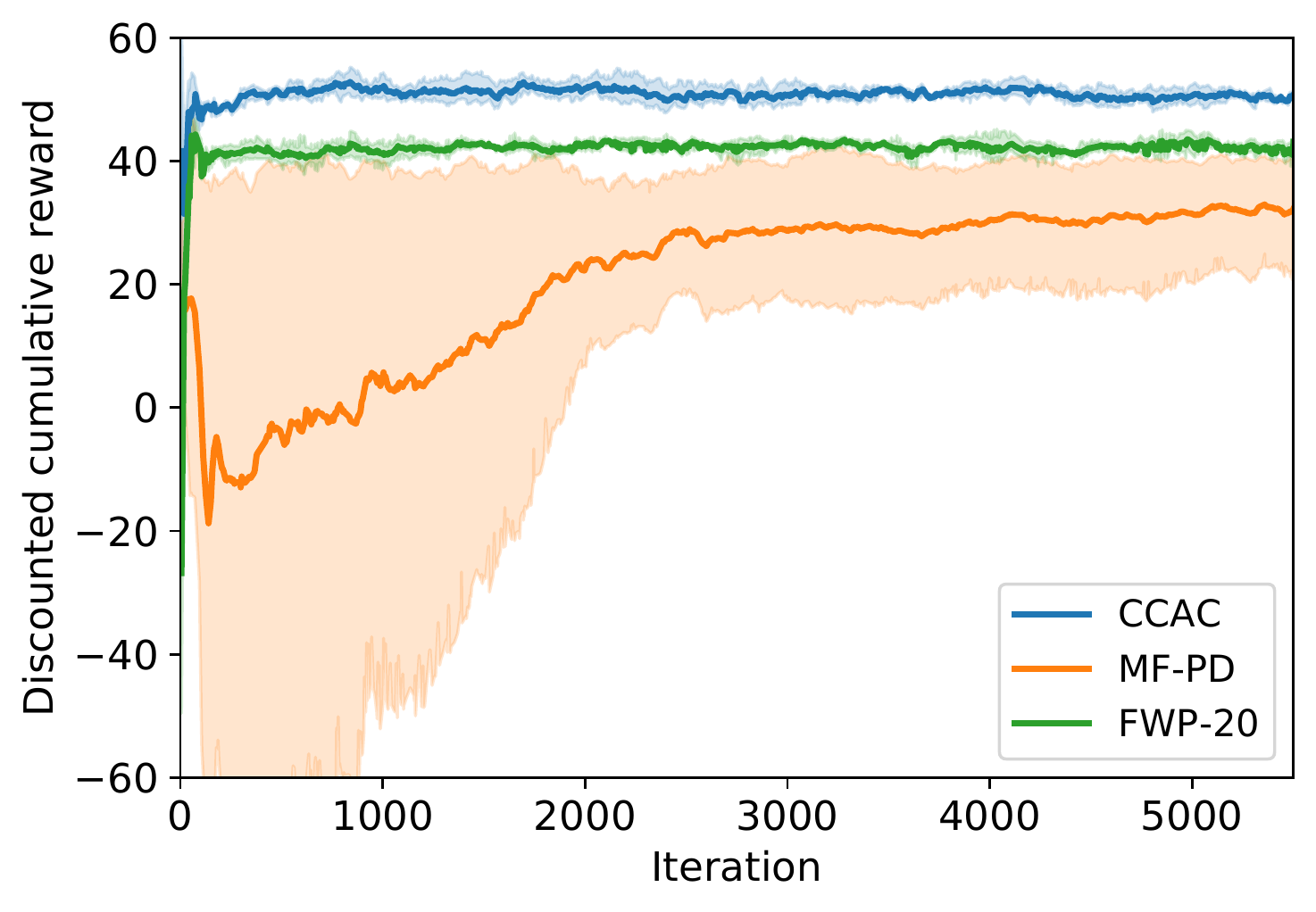}
        \label{fig:return09}
    }
    \subfigure[Discounted cumulative reward under $99.9\%$ threshold]{
	\includegraphics[width=0.38\textwidth]{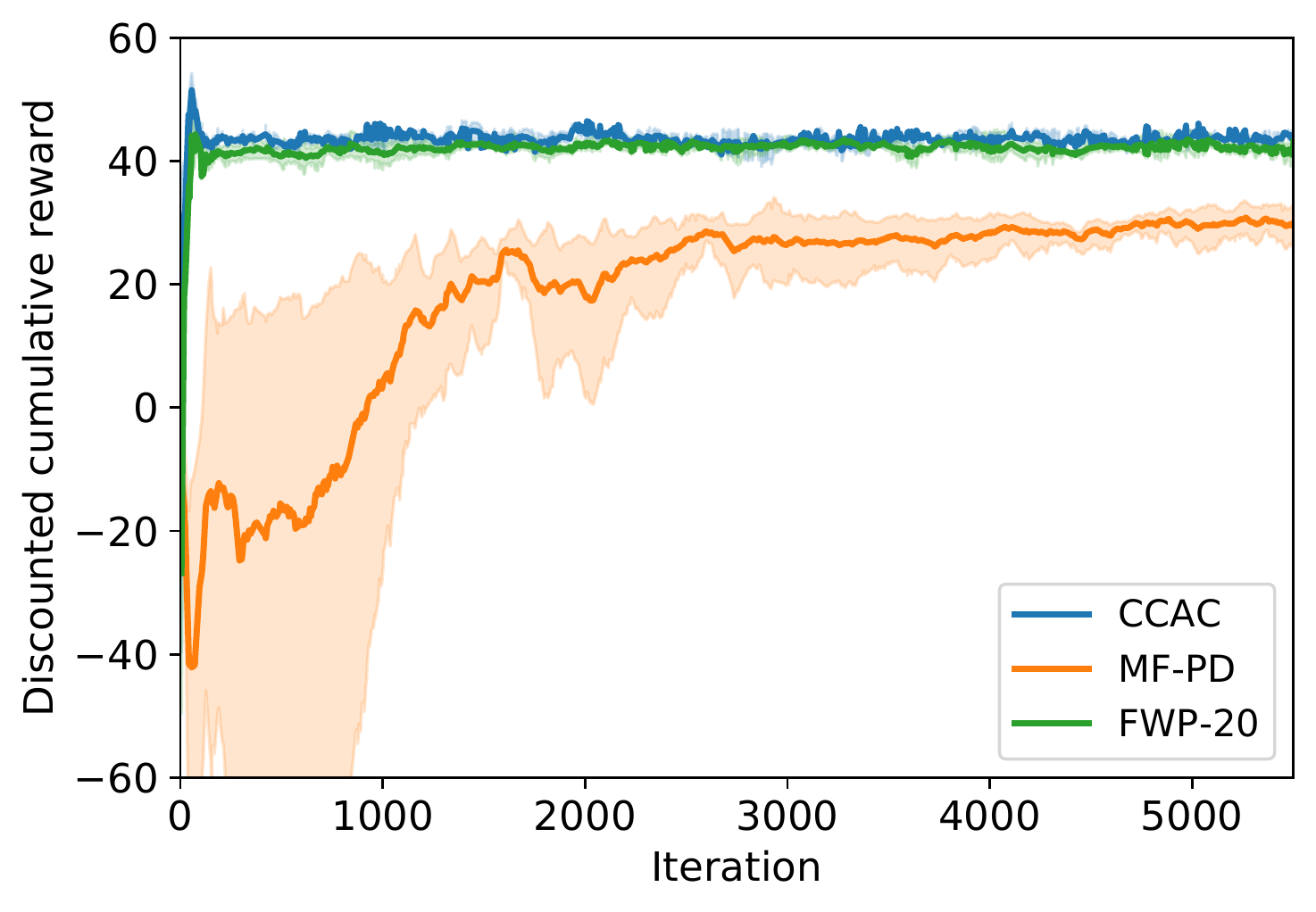}
        \label{fig:return0999}
    }
    \\   
    \subfigure[Safe probability under $90.0\%$ threshold]{
    	\includegraphics[width=0.38\textwidth]{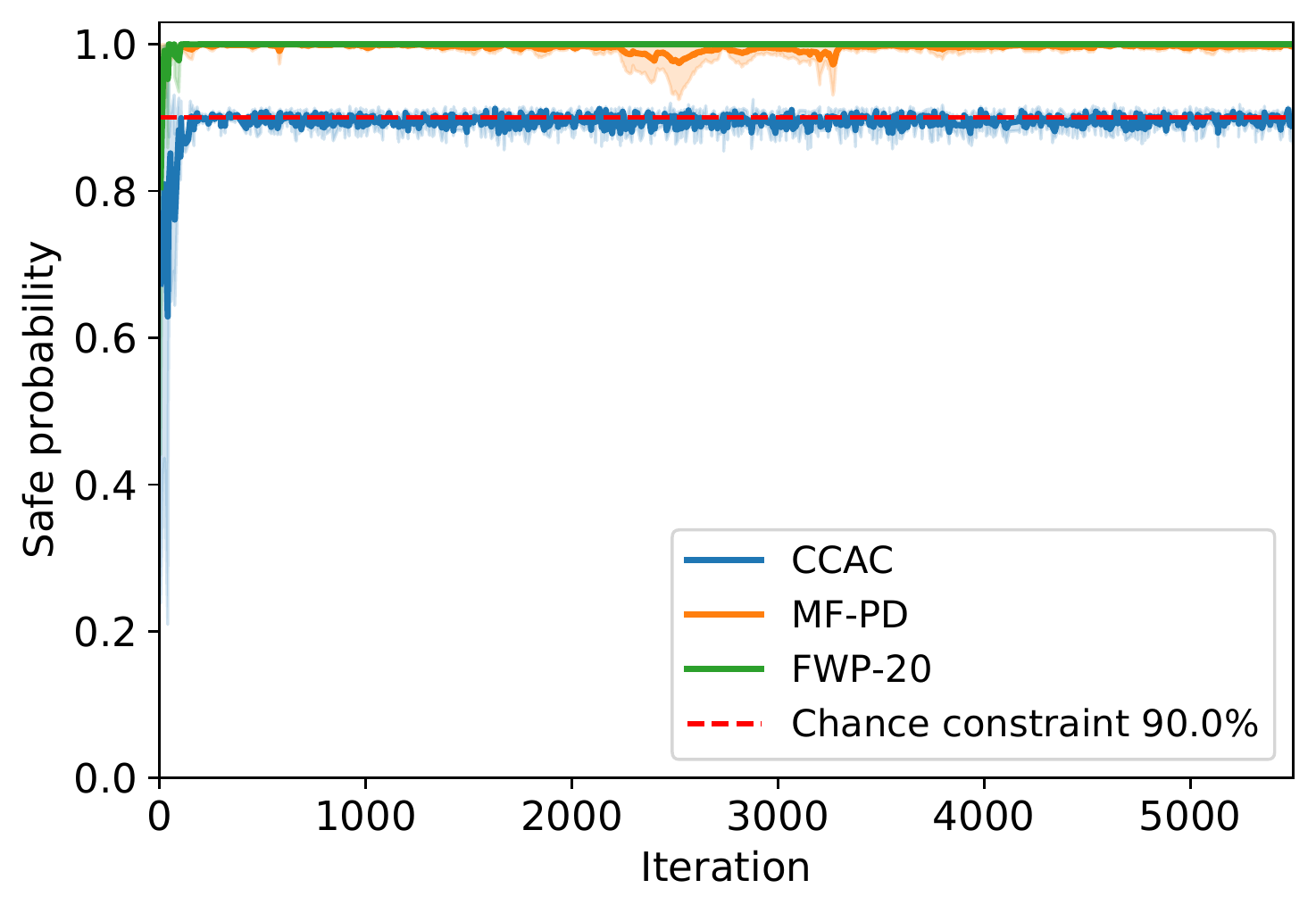}
        \label{fig:safety09}
    }
    \subfigure[Safe probability under $99.9\%$ threshold]{
	\includegraphics[width=0.38\textwidth]{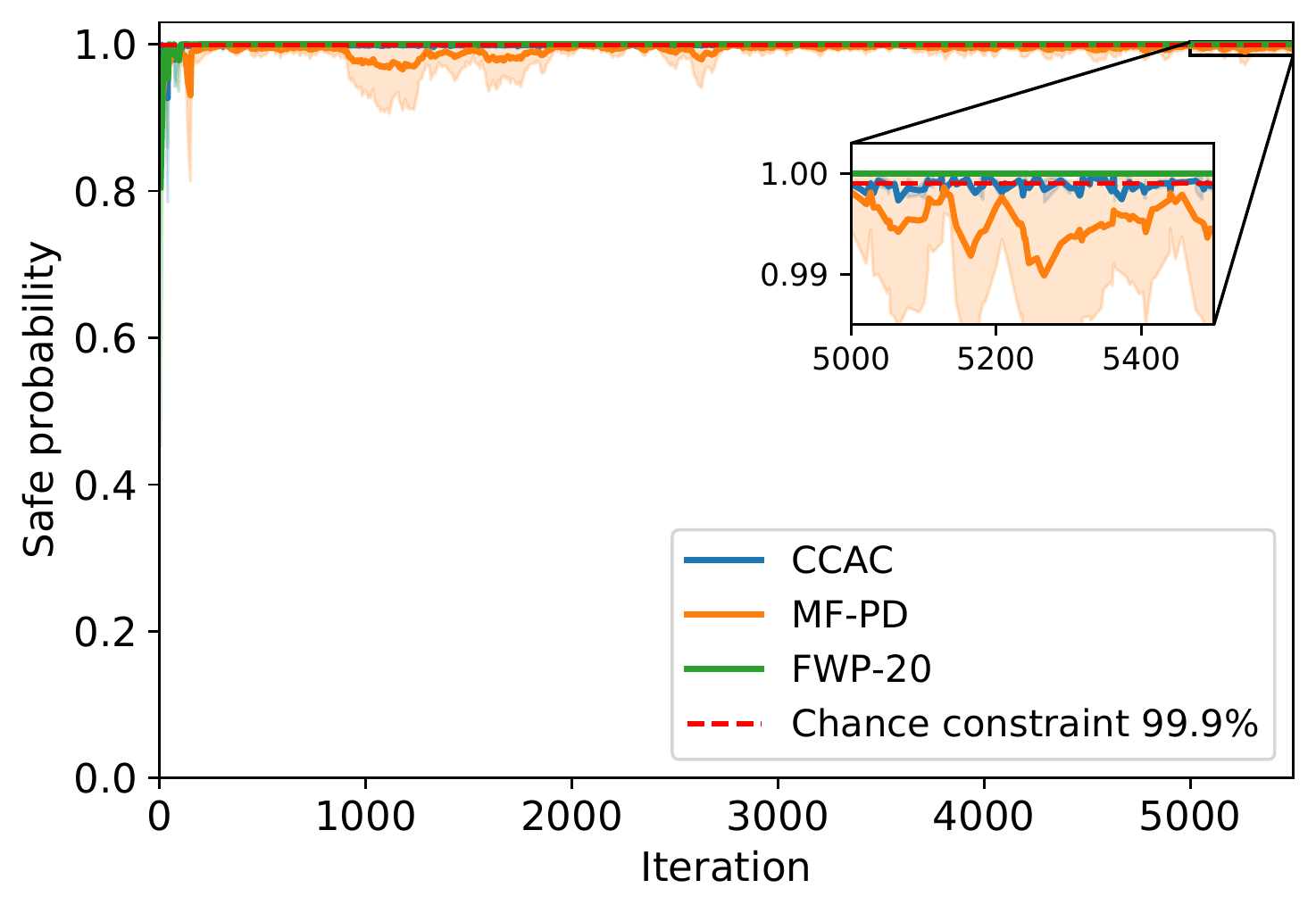}
        \label{fig:safety0999}
    }
    \caption{Comparison of training process among CCAC (chance constrained actor-critic), MF-PD (model-free primal-dual method) and FWP-20 (penalty with fixed weight 20).}
    \label{fig:performance}
\end{figure*}

The system state vector is $s =[v_e \quad v_f \quad \epsilon]^{\top}$ , where $v_e$ denotes the velocity of ego car, $v_f$ is the velocity of front car, and $\epsilon$ is the gap between the two cars. The action $a_t\in (-4,3)$ is the acceleration of ego car, and the disturbance $\xi_t \sim \mathcal{N}(0,1)$ is truncated in the interval $(-5,5)$. $T=0.1$ is the simulation time step. The chance constrained RL problem is defined as
\begin{equation}\begin{aligned}
&\max _{\pi} \mathbb{E}_{s_0,\xi}\left\{\sum_{t=0}^{\infty} \gamma^{t}(0.2v_{e,t}-0.05\epsilon_t)\right\} \\
&\text { s.t. } {\rm Pr}\left\{ \bigcap_{t=1}^{N} \left[\epsilon_{t}>2\right]\right\}\ge1-\delta
\end{aligned}\end{equation}
where $v_{e,t}$ denotes the ego car velocity at time step $t$. In this setting, the agent is expected to drive fast and close to frontal car while keeping a minimum gap of 2m.
\subsection{Implementation Details}
We implement CCAC algorithm on the problem above. Our parameterized actor and critic are both fully-connected neural networks.  Each network has two hidden layers using rectified linear unit (ReLU) as activation functions, with 64 units per layer. The main hyper-parameters of the algorithm are listed in Table \ref{tab:hyper}. The penalty factor in \eqref{exterior_p} is set as $b_k=\text{min}(1000*1.01^k, 10000)$. 
\begin{table}[hbt]
\caption{Hyper-parameters}
\begin{center}
\label{tab:hyper}
\begin{tabular}{lcc}
\hline
Parameters                       & Symbol                & Value    \\ \hline
trajectories number             & $M$                     & 8192      \\
constraint horizon              & $N$                     & 80        \\
discounting factor              & $\gamma $             & 0.98      \\
actor learning rate             & $\alpha_{\theta}$     & 36e-5$\rightarrow$ 2e-5     \\
critic learning rate          & $\alpha_{\omega}$       & 2e-4       \\ 
scale factor                    & $\eta$                & 10         \\
\hline
\end{tabular}
\end{center}
\end{table}
\subsection{Comparison Baselines}
To demonstrate the advantages of CCAC, the performance is compared with model-free primal-dual method (MF-PD) \cite{Paternain2019SafePF}, model-based fixed-weight penalty method (FWP), and two traditional safety techniques, i.e., stochastic model predictive control (SMPC) and safety shielding. The five methods are evaluated under two chance constraint thresholds 90.0\% and 99.9\%, i.e., $\delta=0.1$ and $\delta=0.001$. For FWP method, we test it with different penalty weights in advance and choose a large weight 20 for 99.9\% threshold. Unfortunately, we find it is very hard to select an appropriate weight which produces a 90.0\% safe policy. A small decline of the weight will cause a drastic decline of the safe probability. Therefore, we just choose a weight of 10 although it is relatively unsafe. For simplicity, FWP with weight 10 and 20 are shortly labeled as FWP-10 and FWP-20, respectively. The SMPC method adopts the stochastic tube approach to find the feasible actions under uncertainty \cite{Heirung2018StochasticMP}. The safety shielding method projects the action produced by an unconstrained policy network into the safe action region by solving a constrained optimization problems \cite{dalal2018safe}. To handle the joint chance constraint, Boole's inequality is also used in both SMPC and shielding.  

\subsection{Evaluation Results}
We will analyse the five methods from the aspects of convergence rate, asymptotic performance and computation time.

\begin{table}[hbt]
\caption{Asymptotic performance under 90.0\% Chance Constraint}
\begin{center}
\begin{tabular}{lccc}
\hline
                           & Safe probability & Discounted cumulative reward \\ \hline
CCAC                     & 90.28\%               & 50.17            \\
MF-PD                    & 99.72\%               & 31.84            \\
FWP-10                   & 0.18\%                & 196.10           \\
FWP-20                   & 100.00\%              & 43.07             \\
SMPC                     & 99.41\%               & 43.50             \\
shielding                & 91.84\%               & 42.67            \\  \hline

\end{tabular}
\end{center}
\label{tab:training performance 0.9}
\end{table}

\begin{table}[hbt]
\caption{Asymptotic performance under 99.9\% Chance Constraint}
\begin{center}
\begin{tabular}{lccc}
\hline
       & Safe probability & Discounted cumulative reward          \\ \hline
CCAC                   & 99.88\%               & 44.04            \\
MF-PD                  & 99.43\%               & 29.89            \\
FWP-10                   & 0.18\%                & 196.10           \\
FWP-20                 & 100.00\%              & 43.07             \\
SMPC                   & 99.95\%               & 42.15             \\
shielding              & 91.89\%               & 42.65            \\ \hline
\end{tabular}
\end{center}
\label{tab:training performance 0.999}
\end{table}

\subsubsection{Convergence rate}
For three RL methods (i.e., CCAC, MF-PD and FWP), the learning curves of discounted cumulative reward and joint safe probability in horizon $N$ are plotted in Fig. \ref{fig:performance}. Each curve is averaged over five independent experiments. Besides, the curve of FWP-10 is omitted since it wins unreasonable reward by greatly sacrificing safety. In Fig. \ref{fig:return09}, we notice that model-free algorithm MF-PD converges at about 5000 iterations in terms of reward. Contrarily, the model-based algorithms CCAC and FWP learns at least five times faster and converge within 1000 iterations. Besides, their variances are dramatically lower than MF-PD. These observations confirm the advantage of the utilization of the analytical gradient given by the dynamic model to accelerate and steady the training process. 
\subsubsection{Asymptotic performance}
\label{Asymptotic performance}
The comparisons of asymptotic performance under two chance constraint thresholds are summarized in Table \ref{tab:training performance 0.9} and Table \ref{tab:training performance 0.999}. The proposed CCAC achieves the highest discounted cumulative reward in both constraint thresholds. MF-PD exhibits large conservatism and attains fewer rewards. Especially, MF-PD learns a policy with 99.72\% safe probability even when the required threshold is only 90.00\%. The root cause of conservatism is that MF-PD imposes the chance constraint indirectly by constraining the lower bound of safe probability. On the contrary, CCAC directly solves the original chance constraint and learns an optimal safe policies with less conservatism. The FWP-20 method achieves good performance in terms of both safety and reward in 99.9\% threshold, but FWP-10 fails in 90.0\% threshold.  The traditional safety techniques SMPC and shielding achieve overall good performance, and SMPC is safer than shielding method. But similar to MF-PD, the use of Boole's inequality introduces conservatism, which makes them win fewer rewards than CCAC in 90.0\% threshold. 

Additionally, one may question that the safe probability of CCAC is slightly lower than the threshold in some iterations. We argue that for chance constrained problems, the chance constraint threshold is just a measurement of the safety level, instead of a physical quantity. Therefore, it will not cause huge difference if the safe probability is slightly below the threshold. In our experiment shown in Fig. \ref{fig:safety0999}, the fluctuation range around the constraint threshold is only within 0.5\%. Actually, since the safe probability is estimated by Monte Carlo simulation with a parameterized policy network, such a fluctuation of safe probability is inevitable. 

% In order to demonstrate CCAC's ability to learn polices of different safety level, we test our method with three safe probability: 0.75, 0.9, 0.999. The convergence results are concluded in the table. We stress that the penalty factor is kept constant among three experiments to make sure the results not come from hyper-parameter tuning. 

\begin{figure}[htbp]
\centerline{\includegraphics[width=0.42\textwidth]{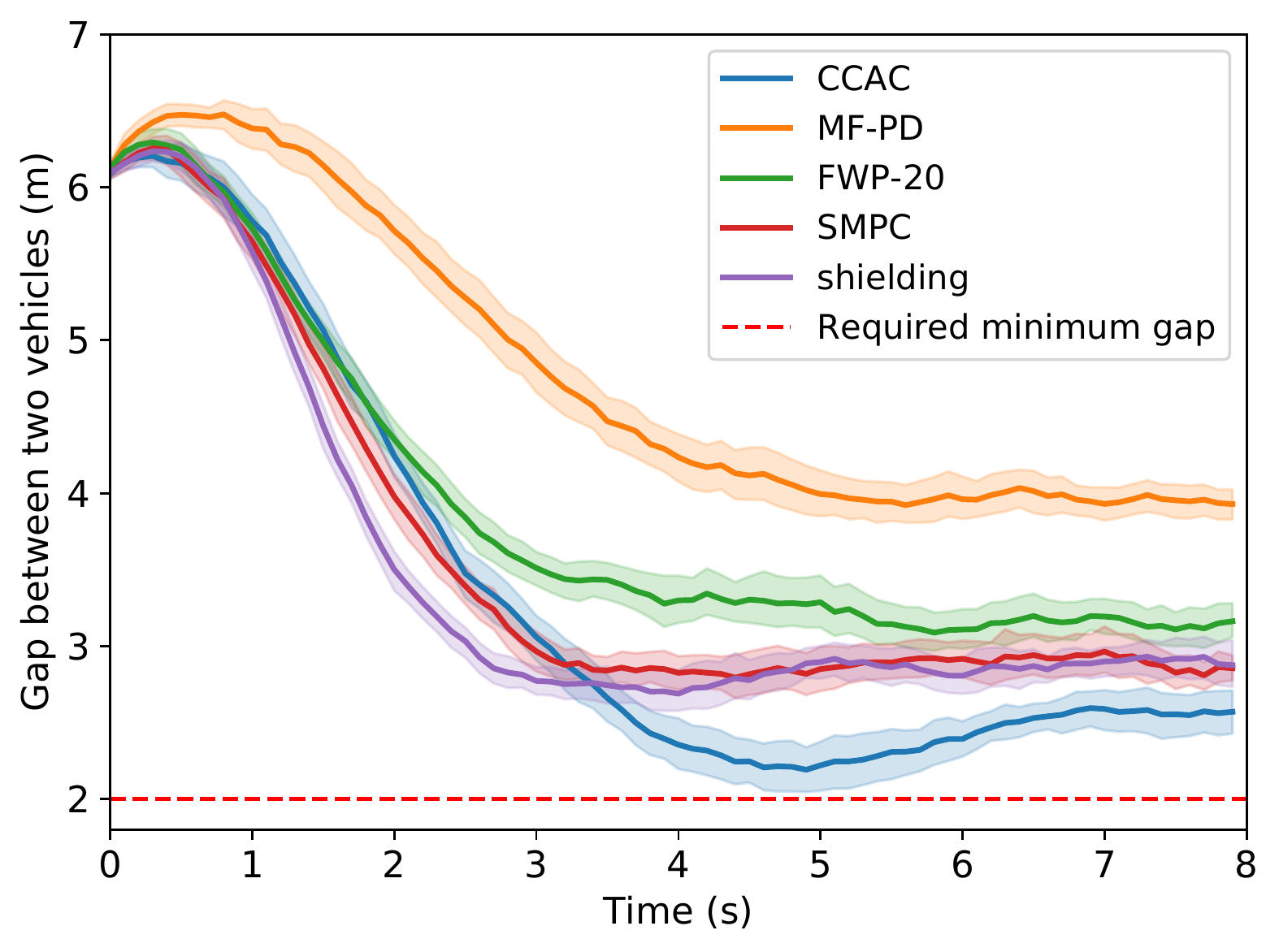}}
\caption{Simulations of car-following gap among CCAC (chance constrained actor-critic), MF-PD (model-free primal-dual), FWP-20 (penalty with fixed weight 20), SMPC (stochastic model predictive control) and shielding.}
\label{fig:gap}
\end{figure}

To give an intuitive comparison of five methods, we implement them under 90.0\% threshold on the same initial state, i.e., $v_e=5, v_f=6$ and $\epsilon=6$. Fig. \ref{fig:gap} demonstrates the curve of car-following gap $\epsilon$ averaged on twenty independent simulations. CCAC keeps the minimum gap while maintaining safety. In contrast, MF-PD is more conservative and retains a large gap from the front car. FWP, SMPC and shielding methods achieve intermediate car-following gaps.

\subsubsection{Online computation efficiency}
In some real-world applications like autonomous driving, online computation efficiency is also essential. The computation efficiency is measured by average one-step computation time, i.e., given a certain state, the average time the controller (or neural networks for three RL methods) takes to compute the action. Table \ref{tab:computation time} summarizes the results of five methods, where the optimization package for SMPC and shielding is Ipopt \cite{Wchter2006OnTI}. Benefitting from the parameterized policy networks, CCAC, MF-PD and FWP achieve a dramatically fast computation speed since they only involve the forward propagation of neural networks. However, traditional safety techniques like SMPC and shielding have to solve the constrained optimization problems online. Thus their computation efficiency is nearly 100 times lower. These results indicate the remarkable advantages of CCAC over traditional safety techniques and confirm the promising potential of CCAC in real-world tasks.   
\begin{table}[H]
\caption{Comparison of one-step computation time}
\begin{center}
\begin{tabular}{lccccc}
\hline
Methods     & CCAC      & MF-PD     & FWP     & SMPC   & shielding                  \\ \hline
time (ms)   & 0.051     & 0.057     & 0.051   & 8.168  & 4.936                      \\ \hline

\end{tabular}
\end{center}
\label{tab:computation time}
\end{table}

In summary, CCAC succeeds in learning a safe but not conservative policy, with a five times faster convergence  rate than existing chance-constrained RL approaches. It also has 100 times higher online computation efficiency than some traditional safety methods.

\section{Conclusion}
\label{sec:Conclusion}
This paper proposed a model-based RL algorithm CCAC applied to safety-critical stochastic systems.
Instead of constraining the lower bound of safe probability like previous methods, CCAC directly solved the original chance constraint and thus avoided conservatism. Besides, CCAC significantly improved the convergence rate by using the gradient of dynamic model. 
The benefits of CCAC were demonstrated in simulations of a stochastic car-following task, where it achieved high reward while satisfying the chance constraint. Additionally, CCAC also had five times faster convergence rate than a model-free method and 100 times higher online computation efficiency than traditional safety methods. 
The application of CCAC to more general environmental dynamics will be investigated in the future.

\bibliographystyle{ieeetr}
\bibliography{ref}

\end{document}